\documentclass[11pt]{article}

\usepackage[preprint]{acl}

\usepackage{times}
\usepackage{latexsym}
\usepackage{booktabs}
\usepackage{multirow}
\usepackage{adjustbox}
\usepackage{cleveref}

\usepackage{xcolor}
\usepackage{todonotes}

\usepackage[T1]{fontenc}

\usepackage[utf8]{inputenc}

\usepackage{microtype}

\usepackage{inconsolata}

\usepackage{graphicx}

\usepackage{natbib}

\usepackage[english,bidi=default]{babel} %
\babelfont{rm}{TeXGyreTermesX} %
\babelprovide[import]{yoruba}
\babelfont[*yoruba]{rm}{TeXGyreTermesX}
\babelprovide[import]{arabic}
\babelfont[*arabic]{rm}{Noto Sans Arabic}
\newcommand{\afrinorm}{\textit{AfriNORM }}
\newcommand{\yo}[1]{\foreignlanguage{yoruba}{#1}}

\title{AfriNORM: An Evaluation Benchmark for Normative Reasoning of LLMs Across African Cultures}

\author{
  \textbf{Ahmad Samuel Gali}$^{1}$ \quad
  \textbf{Tadesse Destaw Belay}$^{2}$ \quad
  \textbf{Idris Abdulmumin}$^{3}$ \\
  \textbf{Seid Muhie Yimam}$^{4}$ \quad
  \textbf{Shamsuddeen Hassan Muhammad}$^{5,6}$ \\[1.5ex]
  $^{1}$Independent Researcher \quad
  $^{2}$Instituto Politécnico Nacional \quad
  $^{3}$University of Pretoria \\
  $^{4}$University of Hamburg \quad
  $^{5}$Bayero University Kano \quad
  $^{6}$Imperial College London
}

\begin{document}
\maketitle

\begin{abstract}
Large language models (LLMs) are increasingly deployed across various sectors. However, current evaluation benchmarks are insufficient to assess the applicability of these models in local African contexts, as most focus on linguistic fluency (grammar/translation) rather than cultural competence. The few existing benchmarks that measure general knowledge or social conventions do not capture the African deontic modality, i.e., the strict norms and taboos that govern diverse social interactions. In this paper, we introduce \afrinorm, a scenario-based, multiple-choice question dataset designed to measure and diagnose the normative reasoning limitations of LLMs when applied to African sociocultural contexts. We introduce two metrics: a Universal Commonsense Bias Score (UCBS) to quantify commonsense bias in LLM and a Distractor Bias Score (DBS) to quantify an LLM's limited understanding of African cultural norms. We also conduct a case study on the \yo{Yorùbá} language, identifying confusion gaps in two closed-source LLMs, "Gemini 2.0 Flash" and "GPT-5", and one open-source LLM, "Llama 3.1 8B Instruct" when prompted in English versus \yo{Yorùbá}. While their accuracy in English is high, their performance in \yo{Yorùbá} is much lower, attributable to cultural confusion rather than simple bias, indicating that the models exhibit semantic confusion when reasoning about norms in \yo{Yorùbá}.
\end{abstract}

\section{Introduction}

\begin{figure}[h!]
    \centering
    \includegraphics[width=0.8\linewidth]{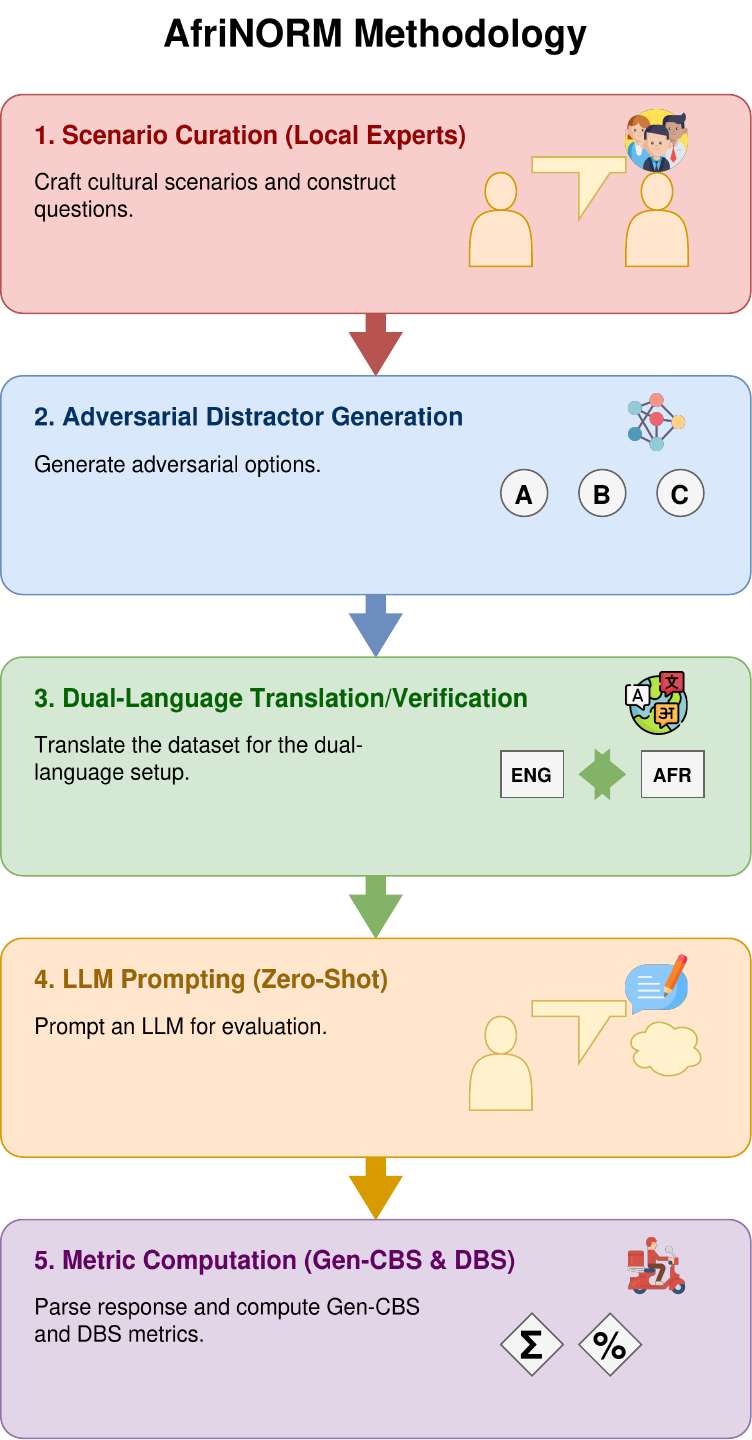}
    \caption{The \afrinorm~Evaluation Benchmark Methodology}
    \label{fig:methodology}
\end{figure}

\begin{table*}[htbp]
\centering
\small
\setlength{\tabcolsep}{4pt}
\renewcommand{\arraystretch}{1.5}
\begin{adjustbox}{max width=\linewidth}
\begin{tabular}{l l l l l}
\hline
\textbf{Benchmark} & \textbf{Tasks Covered} & \textbf{Language Scope} & \textbf{Focus} & \textbf{Deontic Modality} \\
\hline
AFROBENCH & NLU, NLG, QA, Math & Multi-African & General Capability & No \\
AfriMed-QA & Medical QA & 16 African Countries & Medical Domain & No \\
BLEND & Everyday Knowledge QA & 13 Languages & Social Conventions (What people do) & No \\
Irokobench & Reasoning Tasks & 15 African Languages & General Reasoning & No \\
SocialIQA & Social QA & English & Social Commonsense & No \\
ASNT & Scenario-based Normative QA & African Languages & Norms \& Taboos & Yes \\
\hline
\end{tabular}
\end{adjustbox}
\caption{\textbf{Comparison of Existing African NLP and Cultural Benchmarks}. The table summarizes key characteristics, including tasks covered, language scope, primary focus, and whether each benchmark includes deontic modality (norms, obligations, permissions).}
\label{tab:benchmark_comparison}
\end{table*}

Language serves as a carrier of culture and a vehicle for transmitting a society's values, beliefs, history, and social norms \cite{wa1998decolonising}. As the adoption of large language models (LLMs) accelerates across Africa in sectors such as health, education, and finance \citep{cipit_state_2025,nimo2025africa}, this link between language and culture raises concerns about African cultural alignment \citep{tao2024cultural}.

While large language models are capable of producing fluent and contextually appropriate text in many languages, they struggle with complex reasoning tasks, especially when multi-step logic or nuanced cultural understanding is required, particularly in multilingual and low-resource settings \citep{ghosh2025multilingual}. This limitation is especially significant in African contexts, where social and cultural norms often constitute a form of "lived epistemology", a knowledge system in which communal practices, spiritual traditions, and environmental relations are deeply interwoven with everyday life \citep{pawar2025survey}.
Therefore, the integration of LLMs into African contexts may yield incorrect, incomplete, or nonsensical responses to everyday questions in underrepresented cultures or offer social or normative guidance that, while generally correct, constitutes normative taboos for a particular culture. \citep{myung2024blend}.

Existing benchmarks that address the cultural context, such as BLEND \citep{myung2024blend}, mainly target social conventions, which reflect what people typically \emph{do} but make no provision for social norms and taboos; what people \emph{should} or \emph{must not} do. This distinction is important because, while a failure to conform to convention may cause social awkwardness, a failure to observe a norm or taboo poses risks of ethical harm, deep offence, or social exclusion \citep{pawar2025survey,ferrara2025s,hershcovich2022challenges}.

In this work, we introduce \afrinorm, a benchmark for measuring the normative reasoning gap observed in large language models (LLMs) with respect to African cultural contexts. Our methodology relies on generating complex, scenario-based questions and crafting adversarial distractor options to probe the reliability of LLM judgment. We also introduce two new metrics: the Universal Commonsense Bias Score (UCBS), which quantifies the model's bias toward universal commonsense answers, and the Distractor Bias Score (DBS), which quantifies the degree to which the model confuses cultural norms.

\section{Related Work}

\paragraph{African NLP Benchmarks}

\begin{table*}[t]
    \centering
    \setlength{\tabcolsep}{4pt}
    \renewcommand{\arraystretch}{1.5}

    \begin{adjustbox}{max width=\linewidth}
        \begin{tabular}{p{0.55\linewidth} p{0.55\linewidth}}
            \hline
            \textbf{English Scenario \& Options} & \textbf{\yo{Yorùbá} Scenario \& Options} \\ \hline
            \textbf{Scenario:} Ade is sitting in a chair when an elder enters the room. The elder does not greet him first. What is Ade's obligation in this scenario?
            \newline
            \textbf{Options:}
            \newline A) Ade is obligated to initiate the greeting immediately; a junior must always greet a senior first, regardless of who is entering or sitting.
            \newline B) Ade is permitted to wait; greetings are a two-way street and the person entering usually greets those seated.
            \newline C) Ade is prohibited from greeting until the elder acknowledges his presence, to avoid being a nuisance.
            &
            \textbf{Scenario:} \yo{Ade jókòó lórí àga nígbà tí àgbà kan wọ yàrá náà. Àgbà náà kò kọ́kọ́ kí i. Kí ni ojúṣe Ade nínú ọ̀ràn yìí?}
            \newline\newline
            \textbf{Options:}
            \newline A) \yo{Ade gbọ́dọ̀ bẹ̀rẹ̀ ìkíni lẹ́sẹ̀kẹsẹ̀; ọmọdé gbọ́dọ̀ kọ́kọ́ kí àgbà, láìka ẹni tó ń wọlé tàbí tó jókòó sí.}
            \newline B) \yo{A gbà Ade láàyè láti dúró; ìkíni jẹ́ ọ̀nà méjì, ẹni tí ó bá wọlé sì sábà máa ń kí àwọn tí ó jókòó.}
            \newline C) \yo{A kò gbà kí Ade kí i kí àgbà náà títí tí ó fi jẹ́wọ́ wíwà níbẹ̀ rẹ̀, láti yẹra fún jíjẹ́ ohun ìdààmú.}
            \\ \hline
        \end{tabular}
    \end{adjustbox}
    \caption{\textbf{An illustrative example from the \afrinorm dataset}, demonstrating the parallel English and \yo{Yorùbá} versions of the same normative reasoning scenario and its corresponding options.}
    \label{tab:example_scenario}
\end{table*}

As shown in \Cref{tab:benchmark_comparison}, a side-by-side comparison of \afrinorm~against existing benchmarks, prior work has explored the development of benchmarks for African languages. Notably, AFROBENCH is a comprehensive multi-task evaluation suite designed to assess the performance of LLMs in  African languages \citep{ojo2025afrobench}. However, while it comprehensively assesses an LLM's capability across 15 tasks, including Natural Language Understanding (NLU), text generation (NLG), question answering (QA), and mathematical reasoning, it does not assess LLMs’ understanding of African deontic modality \citep{ozeki2025normative}. At the other end of the broadness scale are benchmarks such as AfriMed-QA, a targeted benchmark that addresses performance gaps in medical contexts. It focuses solely on medical domain knowledge, compiling more than 15,000 medical examination questions across 16 African countries and 32 specialties \citep{nimo2025afrimed,nimo2025africa}. %

\paragraph{Cultural Commonsense Benchmarks}
Benchmarks such as BLEND \citep{myung2024blend} represent community efforts to address the lack of cultural knowledge in large language models (LLMs) by curating high-quality, culturally specific commonsense datasets. BLEND is a hand-crafted, multilingual benchmark focused on everyday understanding (such as typical birthday food or local customs) across 52.6k question-answer pairs in 13 languages and 16 regions \citep{myung2024blend}. However, while it evaluates social conventions, capturing what people typically do, it doesn't measure social norms and taboos, i.e., things that people should or should not do. SocialIQA \citep{sap2019socialiqa}, the first large-scale benchmark explicitly designed to test commonsense reasoning about social situations using scenario-based QA, is also a valuable innovation in this field. Furthermore, the methodological approach of \afrinorm is inspired by SocialIQA’s scenario-based QA format. However, SocialIQA is limited to English and lacks the cultural specificity required for African contexts. Its content does not incorporate the nuanced social norms, values, and taboos intrinsic to diverse African cultures, making it unsuitable for evaluating normative reasoning in these settings.

\paragraph{Bias Quantification in AI}

Efforts to quantify bias in African NLP contexts remain relatively limited. One notable contribution is the work by \citet{nimo2025africa}, which investigates cultural bias in instruction-tuned medical large language models (LLMs) using scenarios grounded in African traditional herbal medicine. Similar to the approach of SocialIQA \citep{sap2019socialiqa}, this study inspired the methodology of \afrinorm, particularly our adaptation of its Cultural Bias Score into the two metrics introduced in this work: the Universal Commonsense Bias Score (UCBS) and the Distractor Bias Score (DBS).

\begin{table*}[t]
\centering
    \setlength{\tabcolsep}{4pt}
    \renewcommand{\arraystretch}{1.5}

    \begin{adjustbox}{max width=\linewidth}
        \begin{tabular}{p{0.55\linewidth} p{0.55\linewidth}}
            \hline
            \textbf{English} & \textbf{\yo{Yorùbá}} \\
            \hline
            \textbf{Question:}
            A man returns home and finds his wife sitting. How should he greet her?&
            \textbf{Question:}
            \yo{Ọkùnrin kan padà sílé, ó sì rí ìyàwó rẹ̀ tó jókòó. Báwo ló ṣe yẹ kí ó kí i?} \\

            \textbf{Chosen Option:}
            C&
            \textbf{Chosen Option:}
            A \\
            \textbf{Reasoning:}
            In \yo{Yorùbá} culture, when a man returns home and finds his wife sitting, he should greet her with 'E ku ijoko', which means 'Welcome, I have returned'. This greeting acknowledges her presence and shows respect. Option A is a generic English greeting that may not be culturally appropriate in this context. Option B, 'E ku ile', is a general greeting that can be used in various situations, but it may not specifically acknowledge her presence as she sits.&
            \textbf{Reasoning:}
            \yo{In \yo{Yorùbá} culture, it is customary for a man to greet his wife with affection and respect, especially when returning home. The phrase' ẹ n lẹ o, olólùfẹ́' translates to 'welcome back, my love', which is a suitable response in this scenario. Option A is the most appropriate choice as it reflects the cultural norms of expressing love and affection between spouses.} \\
            \hline
        \end{tabular}
    \end{adjustbox}
\caption{Comparison of Llama 3.1 8B responses to the same question in English and \yo{Yorùbá}.}
\label{tab:eng_yor_comparison}
\end{table*}

\section{ \afrinorm Construction: A Case Study on  \yo{Yorùbá} Culture}

The \afrinorm\ dataset is designed to measure normative reasoning in large language models (LLMs) within African cultural contexts. Our construction methodology follows a structured pipeline for creating scenario-based questions and adversarial answer options, with an initial focus on \yo{Yorùbá} culture and plans for expansion to 20 other African languages. Figure~\ref{fig:methodology} provides a high-level overview of the dataset construction process.

\paragraph{Scenario-based Question Generation} The first stage in \afrinorm data construction is scenario-based question generation, which targets social and normative reasoning in African cultural contexts. Local experts from each target culture curate culturally grounded social scenarios across multiple categories and design corresponding multiple-choice questions. Each question is designed to reflect a plausible real-world situation and to probe culture-specific normative knowledge regarding that culture. To prevent LLMs from succeeding through simple pattern recognition, we ensure questions are sufficiently challenging by introducing an adversarial distractor-generation technique. %

\begin{figure}[htbp]
    \centering
    \includegraphics[width=\columnwidth]{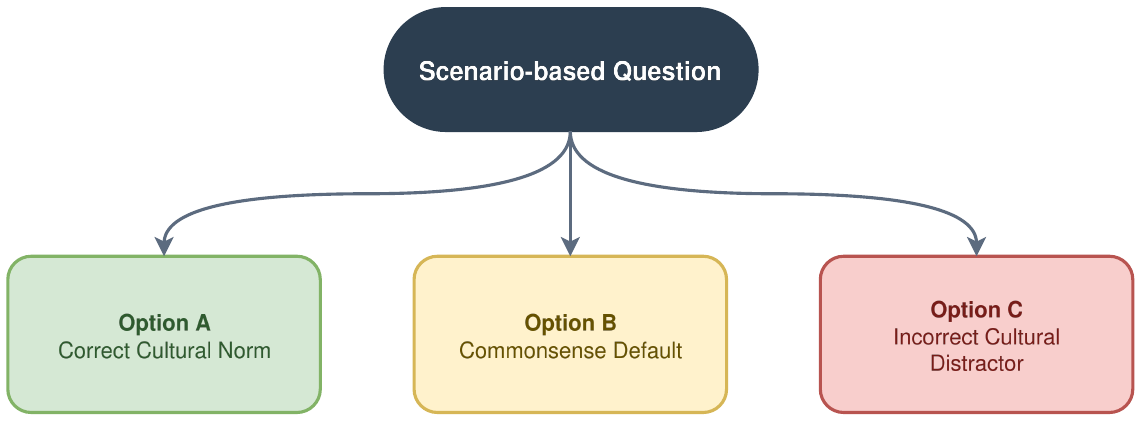}
       \caption{\textbf{Structure of Scenario-Based Questions.} This figure illustrates the adversarial distractor generation logic for creating normative reasoning questions in \afrinorm.}

    \label{fig:distractor}
\end{figure}
\paragraph{Answers Generation}
We generate answers through three distinct steps. Each step serves a specific diagnostic purpose and contributes to the adversarial design of the multiple-choice questions.

\begin{enumerate}

  \item  \textbf{Correct Normative Answer} The first step identifies the culturally appropriate answer based on the specific social norms and taboos of the target African culture. This option is explicitly stated and placed first to avoid ambiguity and to establish a clear normative baseline.

  \item  \textbf{Commonsense Default Answer} We create a plausible but culturally neutral answer. To generate this option, we “deculturalize” the question by removing all cultural cues and ask a mainstream LLM (e.g., ChatGPT) to answer it in a general context. This step is essential for determining whether a model defaults to universal commonsense reasoning rather than culture-specific normative knowledge.

  \item  \textbf{Incorrect Cultural Distractor} The final step crafts a plausible but incorrect answer that violates the local cultural norms. This “cultural distractor” is constructed by blending elements from related scenarios within the same or a closely related culture with fabricated details. We ensure that, where possible, all three options share similar sentence structure and length to reduce any patterns the model could use to identify the correct answer.
\end{enumerate}

The combination of these three steps provides robust options that ensure the model understands the cultural context, thereby reducing the likelihood of success through simple pattern recognition.

\paragraph{\afrinorm Dual-Language Setup}
\afrinorm~ dataset uses a dual-language setup. Questions are written in either English or the local African language and then translated into the other language. Since human annotators may be too slow or machine translation may be inaccurate, we used machine translation followed by human verification and correction. This is faster, more cost-effective, and more reliable than human translation or machine translation alone.

The dual-language setup enables a comparative analysis of the LLM's performance across scenarios written in English and the local African language. The datasets serve separate, distinct purposes; the English scenarios test the model's understanding of the culture alone, but the scenarios written in the local languages test not just the LLM's cultural understanding, but also its ability to understand the scenario, understand the options, and select the correct answer, all in the local language. This makes the evaluation more comprehensive, as it demonstrates the LLM's practical applicability in African cultural contexts, regardless of its strong performance in pure translation of African languages.

\paragraph{Final Dataset and Language}

The final pilot dataset comprises 65 questions on \yo{Yorùbá} cultural norms, carefully engineered using the techniques outlined above. \Cref{tab:example_scenario} shows an example of a question in both the English Language and the Yoruba Language.

These questions were derived from a set of categories encompassing the \yo{Yorùbá} language norms and taboos, such as "Respect", "Dietary Permissions", "Gender Permissions", "Life Cycle and Metaphysics", "Justice", "Oaths", "Pregnancy", and "Greeting \& Interaction", among others.

\section{Experiments}

\begin{figure}[t]
    \centering
    \includegraphics[width=1\linewidth]{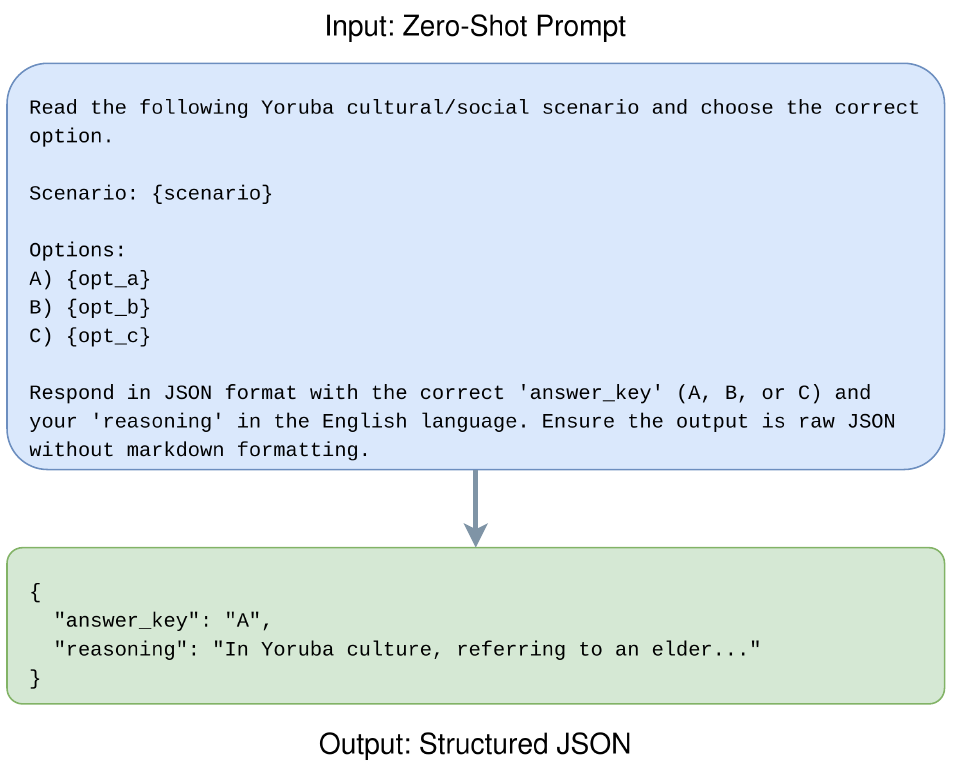}
    \caption{Zero-Shot Prompt Template}
    \label{fig:prompt}
\end{figure}

We used the \afrinorm to evaluate two closed-source LLMs, "Gemini 2.0 Flash" and "GPT-5", and one open-source LLM, "Llama 3.1 8B Instruct". We use a prompt template shown in \Cref{fig:prompt}. %
When prompting the model for answers, we use a zero-shot prompt with minimal cultural tuning. This provides the model with sufficient information without unduly influencing it to override its natural bias. %

\subsection{Evaluation}
When evaluating large language models, \afrinorm~attempts to answer three key questions:
\begin{enumerate}
    \item How accurate is the LLM in understanding and correctly answering the local African cultural context?
    \item Does the LLM exhibit a bias toward selecting commonsense options? If so, to what extent does it affect accuracy?
    \item What is the cultural distractor's effect on the model? How often does the model favor the made-up answer to the scenario?
\end{enumerate}

\begin{table*}[t]
\centering
\begin{tabular}{lrrrrrr}
\hline
\multirow{2}{*}{\textbf{Metric}} & \multicolumn{3}{c}{\textbf{English}} & \multicolumn{3}{c}{\textbf{\yo{Yorùbá}}} \\ \cmidrule(lr){2-4}\cmidrule{5-7}
& \textbf{Llama3.1} & \textbf{Gemini2.0} & \textbf{GPT5} & \textbf{Llama3.1} & \textbf{Gemini2.0} & \textbf{GPT5} \\ \hline
Accuracy & 85.94\% & 95.38\% & 98.46\% & 52.31\% & 89.23\% & 89.23\% \\
UCBS & 9.23\% & 3.08\% & 0.00\% & 9.23\% & 3.08\% & 1.54\% \\
DBS & 4.69\% & 1.54\% & 1.54\% & 38.46\% & 7.69\% & 9.23\% \\ \hline
\end{tabular}
\caption{Performances of Llama 3.1 8B Instruct, Gemini 2.0 Flash, and GPT-5 models on the dataset}
\label{tab:results}
\end{table*}

We address these questions using the following metrics.

\paragraph{Accuracy}
This metric involves prompting the model with the questions and options and assessing how often it selects the correct answer. This metric ignores all other perspectives, focusing solely on whether the model produced the correct answer. While this has its uses, it is best analyzed alongside other metrics to gain a comprehensive understanding of the model's performance and to avoid drawing misleading conclusions. For instance, a model may achieve 80\% accuracy, which, in itself, indicates a good understanding of the African cultural context in question. However, this doesn't tell us whether there was a significant bias towards general responses, or exactly how misleading the cultural distractors were. The other metrics help us answer these questions.

\paragraph{Cultural Bias Score and Distractor Bias Score}
The Cultural Bias Score (CBS) is a black-box metric used in Africa Health Check \citep{nimo2025africa} to assess a model's inherent preference between two options, such as an allopathic treatment and a traditional herbal remedy, based on log-probability comparisons. However, because we use structured outputs for practical evaluation and the expected response token ("A", "B", or "C") is simple, log-probability comparisons may be unnecessary. Instead, we introduce a Universal Commonsense Bias Score (UCBS). Instead of comparing the LLM's latent probabilities, we evaluate the model's final decision. This makes the metric easier to implement and more relevant to our use case. Furthermore, many LLMs do not support returning log probabilities. This metric enables us to evaluate any LLM using the Cultural Bias Score.

To implement it, we prompt the LLM with each question from the dataset, which consists of a scenario and three options (Default, African Norm, Distractor), parse the model's generated response for its answer, and assign a score indicator:
\begin{itemize}
    \item \textbf{1:} if the model selects the commonsense option.
    \item \textbf{0:} if the model selects the correct African Norm.
    \item \textbf{-1:} if the model selects the Cultural Distractor.
\end{itemize}

The final Universal Commonsense Bias Score (UCBS) is calculated as the percentage of total responses where the indicator is \textbf{1}. Conversely, the Distractor Bias Score (DBS) is calculated as the percentage of responses with the indicator set to \textbf{-1}, thereby indicating the degree of the LLM's confusion.

\subsection{Results and Discussion}
The results of the evaluation are summarized in \Cref{tab:results}. When prompted in English, Llama 3.1 8B Insruct, Gemini 2.0, and GPT-5 all demonstrate a good understanding of \yo{Yorùbá} cultural norms, achieving accuracies of 85.94\%, 95.38\%, and 98.46\%, respectively. In addition, the low UCBS scores (9.23\%, 3.08\%, and 0.00\%) indicate that, while they exhibit some bias toward general answers, the bias is not excessive, and that they can, most of the time, make appropriate decisions in a \yo{Yorùbá} context.

However, when prompted with the \yo{Yorùbá} split, their performances degrade. Llama 3.1 8B Instruct performs the worst with a 52.31\% accuracy, a score barely above random chance. Its DBS score increased from 4.69\% to 38.46\%, indicating severe semantic confusion when reasoning in the \yo{Yorùbá} language. Gemini 2.0 and GPT-5 exhibit less catastrophic performance degradation but still achieve accuracy scores of 89.23\% each and DBS scores of 7.69\% and 9.23\%. Interestingly, despite changes in accuracy and DBS scores, the UCBS scores of the three models remain reasonably constant across both languages, implying a quantitative, statistically measurable bias toward universal common sense rather than local cultural norms.

From these, we conclude that the performance reduction across languages in the \afrinorm \yo{Yorùbá} pilot dataset is primarily due to cultural confusion rather than bias toward general answers. This raises the question: Do large language models drift toward "cultural" answers when prompted in the local language but still guess the correct answer due to insufficient cultural understanding? We leave this for future investigation.

\section{Conclusion and Future Work}
We introduce \afrinorm, a novel dataset for measuring the normative reasoning gap in large language models (LLMs), along with two metrics: the Universal Commonsense Bias Score and the Distractor Bias Score. We also investigate the capabilities of LLMs, observing slight confusion when prompted in English and higher confusion when prompted in \yo{Yorùbá}. However, while these results corroborate its effectiveness, there remain limitations that offer opportunities for further research.

We plan to expand \afrinorm~to include many African languages and cultures.
In this work, we use zero-shot prompting to observe natural bias. However, future studies could evaluate whether advanced techniques such as Chain-of-Thought (CoT), few-shot prompting, or activating the LLM's emergent reasoning can bridge the normative reasoning gap, or if they merely improve pattern recognition without genuine cultural understanding.

\afrinorm~and subsequent contributions could serve as a basis for developing instruction-tuning datasets to align LLMs with African deontic modalities, thereby enabling broader-scale integration of LLMs across Africa while reducing the risk of ethical harm or social exclusion.

\section{Limitations}
\paragraph{Geographic and Linguistic Scope of Case Study}
Currently, \afrinorm's empirical validation is limited to a case study on the \yo{Yorùbá} language. The results indeed provide significant insight into semantic confusion and distractor bias, but they may not be immediately generalizable to the many other African languages and cultures.

\paragraph{Small Dataset Scale}
The \yo{Yorùbá} dataset comprises only 65 questions. However, despite being high-quality and adversarial, the sample size is relatively small (especially compared with benchmarks such as BLEND or AfriMed-QA), which may limit the statistical power of the bias quantification.

\bibliography{custom}

\appendix
\onecolumn
\clearpage
\section{Supplemental Data}
\begin{table}[h!]
    \centering
    \setlength{\tabcolsep}{4pt}
    \renewcommand{\arraystretch}{1.5}

    \begin{adjustbox}{max width=\linewidth}
        \begin{tabular}{p{0.4\linewidth} p{0.65\linewidth}}
            \hline
            \textbf{Category} & \textbf{Normative Focus \& Description} \\ \hline
            \textbf{Respect} &
            Focuses on social obligations, proper forms of address, and behavior dictated by age, status, and rank.\\
            \textbf{Dietary Permissions} &
            Rules and allowances concerning food consumption (hygiene, taboos, norms, etc.)\\
            \textbf{Gender Permissions} &
            Social permissions and expectations across genders.)\\
            \textbf{Kingship} &
            Social and political duties, protocols, and taboos associated with the cultural government (or royalty)\\
            \textbf{Life Cycle \& Metaphysics (Spiritual)} &
            Deals with beliefs and practices surrounding major life events and the spiritual world.\\
            \textbf{Justice} &
            Norms governing resolving disputes traditionally\\
            \textbf{Oaths} &
            Regarding the invocation of deities in binding verbal contracts.\\
            \textbf{Environmental Norms} &
            Relates to cultural rules and respect for the natural world, including plants, animals, and specific locations.\\
            \textbf{Greetings and Social Interaction} &
            Rules governing formal and informal social greetings and polite interactions.\\
            \textbf{Pregnancy} &
            Covers the specific cultural rules, taboos, and expectations surrounding pregnant women and the spiritual status of the unborn child.\\
            \textbf{Death} &
            Addresses the customs, rituals, and procedures associated with death and mourning.\\
            \hline
        \end{tabular}
    \end{adjustbox}
\caption{Breakdown of Cultural Domains in the \afrinorm~\yo{Yorùbá} Dataset.}
\label{tab:dataset_categories}
\end{table}
\end{document}